%% file: main.tex
\documentclass[conference]{IEEEtran}
\IEEEoverridecommandlockouts
\usepackage{graphicx}
\usepackage{amsmath}
\usepackage{outlines}
\usepackage{ifthen}
\usepackage{bm}
\usepackage{amsfonts}
\usepackage{textcomp}
\usepackage{braket}
\usepackage{mathtools}
\usepackage{tabularx}
\usepackage{multirow}
\usepackage[inline,shortlabels]{enumitem}
\usepackage{booktabs}
\usepackage{algorithm}
\usepackage{algpseudocode}
\usepackage{caption}
\usepackage{authblk}
\usepackage{color}
\usepackage{subcaption}
\usepackage{tikz}
\usepackage{tablefootnote}
\usepackage{multicol,multirow}
\usetikzlibrary{fit,backgrounds,arrows,decorations.markings,arrows.meta}

\tikzset{
    vertex/.style = {
        circle,
        fill            = black,
        outer sep = 2pt,
        inner sep = 1pt,
    }
}

\def\BibTeX{{\rm B\kern-.05em{\sc i\kern-.025em b}\kern-.08em
    T\kern-.1667em\lower.7ex\hbox{E}\kern-.125emX}}

\newcolumntype{L}[1]{>{\raggedright\arraybackslash}p{#1}}
\newcolumntype{C}[1]{>{\centering\arraybackslash}p{#1}}
\newcolumntype{R}[1]{>{\raggedleft\arraybackslash}p{#1}}

\DeclareMathOperator*{\argmin}{arg\,min}
\newcommand{\spec}{{\fontfamily{lmtt}\selectfont SPEC\textsuperscript{2}}}
\newcommand{\norm}[1]{\left\lVert #1 \right\rVert}

\newcommand{\paren}[1]{\left( #1 \right)}
\newcommand{\Xin}[1][]{
    \ifthenelse{\equal{#1}{}}
    {\bm{X}}
    {\bm{X}^{#1}}}
\newcommand{\Yout}[1][]{
    \ifthenelse{\equal{#1}{}}
    {\bm{Y}}
    {\bm{Y}^{#1}}}
\newcommand{\W}[1][]{
    \ifthenelse{\equal{#1}{}}
    {\bm{W}}
    {\bm{W}^{#1}}}
\newcommand{\Ufreq}[1][]{
    \ifthenelse{\equal{#1}{}}
    {\widetilde{\bm{U}}}
    {\widetilde{\bm{U}}^{#1}}}
\newcommand{\Wfreq}[1][]{
    \ifthenelse{\equal{#1}{}}
    {\widetilde{\bm{W}}}
    {\widetilde{\bm{W}}^{#1}}}
\newcommand{\Zfreq}[1][]{
    \ifthenelse{\equal{#1}{}}
    {\widetilde{\bm{Z}}}
    {\widetilde{\bm{Z}}^{#1}}}
\newcommand{\Xfreq}[1][]{
    \ifthenelse{\equal{#1}{}}
    {\widetilde{\bm{X}}}
    {\widetilde{\bm{X}}^{#1}}}
\newcommand{\FFT}[1][]{
    \ifthenelse{\equal{#1}{}}
    {\mathcal{F}}
    {\mathcal{F}_{#1}}}
\newcommand{\IFFT}[1][]{
    \ifthenelse{\equal{#1}{}}
    {\mathcal{F}^{-1}}
    {\mathcal{F}_{#1}^{-1}}}
\newcommand{\fin}{c_\text{in}}
\newcommand{\fout}{c_\text{out}}
\newcommand{\ha}{h_\text{act}}

\newcommand{\hk}{h_\text{krn}}

\newcommand{\nnz}{\text{\fontfamily{lmtt}\selectfont NNZ}}
\newcommand{\sz}{\text{\fontfamily{lmtt}\selectfont Size}}
\newcommand{\loss}{\text{\fontfamily{lmtt}\selectfont Loss}}
\newcommand{\ind}{\text{\fontfamily{lmtt}\selectfont Ind}}
\newcommand{\pnty}{\text{\fontfamily{lmtt}\selectfont Pty}}

\DeclarePairedDelimiter\ceil{\lceil}{\rceil}

\begin{document}

\title{\spec: \underline{SPEC}tral \underline{SP}ars\underline{E} \underline{C}NN Accelerator on FPGAs\\
}

\author[1]{\small{Yue Niu}}
\author[1]{Hanqing Zeng}
\author[1]{Ajitesh Srivastava}
\author[1]{Kartik Lakhotia}
\author[2]{Rajgopal Kannan}
\author[3]{Yanzhi Wang}
\author[1]{Viktor Prasanna}
\affil[1]{University of Southern California\\
\{yueniu,zengh,ajiteshs,klakhoti,prasanna\}@usc.edu}
\affil[2]{US Army Research Lab-West\\
rajgopal.kannan.civ@mail.mil}
\affil[3]{Northeastern University\\yanz.wang@northeastern.edu}


\maketitle

\begin{abstract}
To accelerate inference of Convolutional Neural Networks (CNNs), various techniques have been proposed to reduce computation redundancy.
Converting convolutional layers into frequency domain significantly reduces the computation complexity of the sliding window operations in space domain. 
On the other hand,
weight pruning techniques address the redundancy in model parameters by converting dense convolutional kernels into sparse ones. 
To obtain high-throughput FPGA implementation, we propose {\spec} -- the first work to \emph{prune and accelerate spectral CNNs}. 
First, we propose a systematic pruning algorithm based on Alternative Direction Method of Multipliers (ADMM). The offline pruning iteratively sets the majority of spectral weights to zero, without using any handcrafted heuristics. Then, we design an optimized pipeline architecture on FPGA that has efficient random access into the sparse kernels and exploits various dimensions of parallelism in convolutional layers.
Overall, {\spec} achieves high inference throughput with extremely low computation complexity and negligible accuracy degradation. 
We demonstrate {\spec} by pruning and implementing LeNet and VGG16 on the Xilinx Virtex platform. After pruning $75\%$ of the spectral weights, {\spec} achieves $0\%$ accuracy loss for LeNet, and $<1\%$ accuracy loss for VGG16. 
The resulting accelerators achieve up to $24\times$ higher throughput, compared with the state-of-the-art FPGA implementations for VGG16. 
\end{abstract}


\input{1_into.tex}
\input{2_background.tex}
\input{3_pruning.tex}

\input{4_arch.tex}
\input{5_exp.tex}
\input{6_conclusion.tex}
\input{7_ack.tex}

\bibliographystyle{IEEEtran}
\bibliography{cite}

\end{document}

%% file: 1_into.tex
\section{Introduction}
\label{sec: intro}

Convolutional Neural Networks (CNNs) are powerful deep learning models, widely deployed for computer vision tasks \cite{AlexNet,VGG16}. 
The high accuracy of CNNs comes with the price of high computation cost. For example, VGG16 \cite{VGG16} requires more than 30 billion multiplications and additions to inference a single image. 
Although light-weight CNN models \cite{mobilenet_v1,shufflenet_v1} have been recently proposed to meet the hardware constraints of embedded systems, large models \cite{VGG16,ResNet} are still irreplaceable in many advanced computer vision tasks. Motivated by the computation challenge, there have been many attempts to speedup CNN inference. 
From the algorithm perspective, spectral convolution \cite{Hanqing_FPGA18} and spatial weight pruning \cite{han2015learning} are two successful methods to reduce the number of operations of a CNN. 
From the architecture perspective, hardware designs have been tailored on FPGAs to fit the computation and communication characteristics of inference. 

Although accelerators co-designed with the algorithmic and architectural innovations have shown impressive throughput and latency performance, challenges still exist in state-of-the-art approaches. 
On the one hand, while spectral CNN accelerators \cite{Chi_FPGA17,Hanqing_FPGA18} perform exact inference and achieve $3\times$ to $4\times$ higher throughput than spatial accelerators, they have significantly higher memory requirement due to enlarged spectral kernels. 
On the other hand, weight pruning on CNNs has been explored to reduce computation and storage requirements simultaneously. However, no spectral domain pruning has been performed in the literature. Further, many pruning approaches (e.g., \cite{han2015learning,guo2016dynamic,wen2016learning, Lowrank_IEEE_2019}), depend on heuristics, making the quality of pruning (in terms of accuracy) sensitive to training data. 
In addition, the irregular sparse structure of the pruned kernels may result in low hardware efficiency when performing the sliding window operation of spatial convolution. Taking the hardware overhead into account, a heavily pruned spatial CNN may not lead to significantly faster inference. Not only in CNN does sparse operations need particular optimization but also in other applications \cite{zhou2017design}, \cite{bader2007design}.

To overcome the above challenges, we propose \spec: a novel approach to speedup CNN inference by applying systematic weight pruning to spectral convolutional layers.
Inference of {\spec} has extremely low computation complexity since (i) the spectral transformation reduces computation redundancy, and (ii) the kernel pruning reduces parameter redundancy.
Memory requirement of {\spec} is also significantly reduced, since after pruning, we only store on chip the non-zeros of the sparse kernels.
Finally, unlike other implementations based on spatial pruning, 
{\spec} achieves high hardware efficiency. 
{\spec} replaces the sliding window operation (of spatial convolution) with element-wise multiplications (of spectral convolution), and enables us to develop a sparse tensor computation engine achieving close to $100\%$ DSP utilization. 
Overall, the systematic pruning algorithm of {\spec} ensures both high accuracy and high pruning rate of spectral CNNs, and the corresponding FPGA design achieves high inference throughput.
The main contributions of this paper are:
\begin{outline}
\1 We propose {\spec} to achieve high throughput inference of spectral sparse CNNs, by:
    \2 \emph{Spectral weight pruning formulation} using Alternative Direction Method of Multipliers (ADMM) algorithm. It minimizes classification error while constraining the number of non-zero weights based on target sparsity.
    \2 \emph{Spectral retraining} of the model to iteratively fine-tune non-zero spectral weights and improve accuracy. 
    \2 \emph{Sparse convolution accelerator}, which implements the pruned spectral convolution on FPGA. The accelerator exploits various dimensions of parallelism, and uses small number of data replications to significantly reduce BRAM conflicts and pipeline stalls. 
\1 We extensively evaluate {\spec} on the below metrics:
    \2 \emph{Inference accuracy}: We show that pruning $75\%$ weights of the spectral kernels leads to negligible accuracy loss ($0\%$ for LeNet on MNIST dataset and $<1\%$ for VGG16 on CIFAR-10 dataset).
    \2 \emph{Inference throughput}: We implement the accelerator on Xilinx Virtex-7 XC7VX690T FPGA. The design achieves $99\%$ of the theoretically optimal throughput performance and $24\times$ higher throughput than state-of-the-art.
\end{outline}

%% file: 2_background.tex
\section{Background and Related Work}
\label{sec: back}


\subsection{Spectral CNNs}
\label{sec: back spec}

The main building blocks of a CNN are convolutional layers. 
A convolutional layer convolves the layer inputs with the kernels to generate the layer outputs (i.e., activations). Let  
$\Xin\in \mathbb{R}^{b\times \fin\times\ha\times\ha}$ and $\Yout\in\mathbb{R}^{b\times \fout\times\ha'\times\ha'}$ be the input and output tensors of a convolutional layer, where $b$ denotes the batch size, $\fin$ and $\fout$ denote the number of input and output channels respectively, and $\ha$ and $\ha'$ denote the spatial dimensions (width and height, respectively). 
Let $\W\in\mathbb{R}^{\fout\times\fin\times\hk\times\hk}$ be the tensor of spatial convolutional kernels,
where $\hk$ denotes the spatial dimension (width and height). 
We use $\FFT_{n}\paren{\cdot}$ and $\IFFT_n\paren{\cdot}$ for the $n\times n$ 2D Fast Fourier Transform (FFT) operation and its inverse. Let ``\texttildelow'' denote the corresponding spectral domain variable. For example, let $\Wfreq_{j,i}=\FFT_{\ha+\hk-1}\paren{\W_{j,i}}$ and $\Xfreq_{k,i}=\FFT_{\ha+\hk-1}\paren{\Xin_{k,i}}$. 
A convolutional layer operates as below:

\begin{equation}
\label{eq: spectral conv}
    \Yout_{k,j} = \underbrace{\sum\limits_{i=1}^{\fin}\Xin_{k,i} * \W_{j,i}}_{\text{Spatial convolution}} = \underbrace{\IFFT_{\ha+\hk-1}\paren{\sum\limits_{i=1}^{\fin}\Xfreq_{k,i}\circ \Wfreq_{j,i}}}_{\text{Spectral convolution}}
\end{equation}

\noindent where $1\leq k\leq b$ and $1\leq j\leq \fout$; ``$*$'' denotes the sliding window operation of spatial convolution; ``$\circ$'' denotes the Hadamard product operation (element-wise multiplication) of spectral convolution. Note that Equation \ref{eq: spectral conv} applies to convolution layers of stride 1 and padding $\hk-1$. 
For padding less than $\hk-1$, we will further crop the $\Yout_{k,j}$. For stride larger than 1, we will further slice the $\Yout_{k,j}$.\footnote{When stride is larger than 1, spectral convolution may not lead to reduced computation complexity. However, most of the convolutional layers of modern CNNs (e.g., \cite{VGG16,ResNet,googlenet}) have stride equal to 1.} 

We apply the Overlap-and-Add (OaA) technique \cite{Chi_FPGA17}, \cite{Hanqing_FPGA18}. 
To perform spectral convolution with $n\times n$ FFT (where $\hk \leq n\leq \ha+\hk-1$), we follow the below steps for each batch index $k$, and channel indices $i$, $j$:

\begin{enumerate}
    \item Partition $\Xin_{k,i}$ into tiles of $\Xin^{t,s}_{k,i}\in \mathbb{R}^{m\times m}$, where $m=n-\hk+1$ and $1\leq t,s\leq \ceil{\frac{\ha}{m}}$;
    \item Transform the tiles $\Xin^{t,s}_{k,i}$ and kernels $\W_{j,i}$ into spectral representation (after proper padding), as $\Xfreq^{t,s}_{k,i} = \FFT_n\paren{\Xin^{t,s}_{k,i}}$ and $\Wfreq_{j,i} = \FFT_n\paren{\W_{j,i}}$;
    \item Obtain output tiles $\Yout^{t,s}_{k,j} = \IFFT_n\paren{\sum\limits_{i=1}^{\fin}\Xfreq^{t,s}_{k,i}\circ \Wfreq_{j,i}}$;
    \item Combine $\Yout^{t,s}_{k,j}$ into $\Yout_{k,j}$ by OaA. For each tile $\Yout^{t,s}_{k,j}$, place its adjacent tiles (i.e., $\Yout_{k,j}^{t\pm 1,s}$, $\Yout_{k,j}^{t,s\pm 1}$) with $\hk-1$ horizontally or vertically overlapping pixels. We obtain the final $\Yout_{k,j}$ by adding the overlapped pixels.
\end{enumerate}

With OaA, we can select the FFT size $n$ to balance the computation and communication capacity of the target FPGA. Therefore, inference throughput can be maximized. 
It has been shown in \cite{Hanqing_FPGA18} that, setting $n=8$ or $16$ reduces the computation complexity of spatial convolution by $3\times$ to $4\times$, and improves the overall FPGA throughput by up to $5\times$. 

While state-of-the-art accelerator \cite{Hanqing_FPGA18} already achieves high throughput, such design only applies to unpruned spectral CNNs.
In this paper, we significantly improve the design of \cite{Hanqing_FPGA18} by reducing the redundancy in model parameters $\Wfreq$. 

\subsection{CNN Weight Pruning}

Pruning algorithms leverage redundancy in CNN weight parameters. 
For a CNN with convolutional layers and fully connected layers, in general, convolutional layers contribute to most ($99\%$ for VGG16) of the computation workload, but contain only a small portion ($10.6\%$ for VGG16) of the model weights. 
Thus, compared with fully connected layers, pruning convolutional layers is more important, yet much more difficult as well. 
For spatial CNNs, pruning rate on fully connected layers can be as high as $30\times$ \cite{zhang2018systematic,han2015deep}, while pruning rate on convolutional layers is often less than $10\times$ \cite{zhang2018systematic}.

Below we summarize state-of-the-art works in pruning spatial CNNs. 
The pioneer work \cite{han2015learning} uses an iterative heuristic to prune  weights of small magnitudes and achieves $2.7\times$ reduction in convolutional weights on AlexNet \cite{AlexNet}. 
This method has been extended in two directions.
The first is to improve compression rate by using more sophisticated heuristics.
For example,
\cite{guo2016dynamic} incorporates both weight pruning and growing, and \cite{wen2016learning} uses $L_1$ regularization. 
Pruning of \cite{dai2017nest} is based on a genetic algorithm. 
The second direction is to enhance the hardware implementation efficiency by  deriving an effective tradeoff between accuracy and pruning rate, e.g., the energy-aware pruning \cite{yang2017designing}, and structure-aware pruning \cite{he2017channel,wen2016learning}. FPGA hardware accelerators \cite{wang2018c,zhao2018building} have also been investigated to accommodate pruned CNNs, by leveraging the reconfigurability in on-chip resources. Recently, the authors of \cite{zhang2018systematic} have developed a systematic weight pruning framework  based on the powerful optimization tool ADMM (Alternating Direction Method of Multipliers) \cite{boyd2011distributed}. 
Such framework consistently achieves higher pruning rate than prior arts.

Besides the works on spatial CNN pruning, \cite{liu2018frequency} proposes a frequency-domain compression technique which first converts inputs and kernels of all layers into frequency domain using discrete cosine transform (DCT), then dynamically prunes unimportant connections. Although this method reduces parameters in convolutional layers by up to $70\%$, it is difficult to leverage this technique in hardware platform. 

With performance brought by spectral convolution algorithm \cite{Hanqing_FPGA18}, combining it with pruning creates more potential for high throughput hardware implementation. 
To the best of our knowledge, {\spec} is the first work to exploit such opportunity.

\subsection{CNN Accelerators}
By leveraging the computation and memory access patterns of CNNs,
many domain-specific architectures have been proposed for accelerating inference \cite{Lowrank_GlobalSIP_2017, Caffein, FPGA2019Xilinx,ISCA2017SCNN,Chi_FPGA17,Hanqing_FPGA18}. Among these works, \cite{Caffein} shows a unified convolutional matrix-multiplication representation for both convolutional and fully-connected layers. This micro-architecture is optimized for both computation speed and resource utilization. In \cite{FPGA2019Xilinx}, a pipeline and DRAM-free architecture is proposed to simplify data movement between memory and processing elements. This design achieves very high working frequency as well as DSP utilization. On the other hand, to exploit sparsity in pruned CNNs, \cite{ISCA2017SCNN} introduces a sparse computation architecture aware of zeros in both weights and activations. 
Beside acceleration in spatial domain, \cite{Chi_FPGA17} and \cite{Hanqing_FPGA18} accelerate convolutional layers in spectral domain, where the computation primitives are Fast Fourier Transforms and Hadamard products. 
Using the additional Overlap-and-Add technique, computation complexity (of unpruned CNNs) can be significantly reduced without hardware reconfiguration. 
ASICs also demonstrate promising performance in processing CNNs due to their customization ability and low power consumption. Google TPUs \cite{GoogleTPU} employ systolic array based matrix multiplication units to execute convolutional and fully-connected layers with $70\times$ higher power efficiency than GPUs. 



%% file: 3_pruning.tex
\section{Pruning Spectral CNNs}
\label{sec: pruning}


\subsection{Overview}

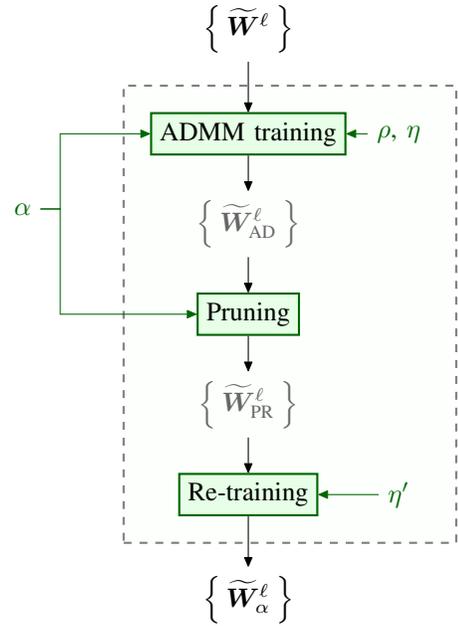
\begin{figure}[!htb]
    \centering
    \input{./fig/overview_prune.tex}
\caption{{\spec} spectral pruning overview}
    \label{fig: prune overview}
\end{figure}

The goal of weight pruning is to convert the dense spectral kernels $\Wfreq$ into sparse ones, while keeping the overall classification accuracy of CNNs unaffected.
ADMM is an optimization framework that we can utilize to iteratively achieve the above goal, without using handcrafted heuristics. 

As summarized by Figure \ref{fig: prune overview}, the inputs to our pruning algorithm are the $L$-layer spectral CNN with its unpruned weights $\Set{\Wfreq[\ell]|1\leq \ell \leq L}$, and the target pruning factor $\alpha$. 
The output is the CNN with sparse spectral weights $\Set{\Wfreq^{\ell}_\alpha}$, where the subscript $\alpha$ ($>1$) means that at most $\frac{1}{\alpha}$ of the tensor elements are non-zeros.
Spectral pruning of {\spec} consists of three steps. 
The \emph{ADMM training} step rescales the weights by joint consideration of the classification accuracy and sparsity requirements. 
The outputs $\Set{\Wfreq[\ell]_\text{AD}}$ are still dense tensors, where 
$\paren{1-\frac{1}{\alpha}}$ portion of the elements are driven to close-to-zero values.
The \emph{pruning} step simply removes the close-to-zero values of $\Set{\Wfreq[\ell]_\text{AD}}$ and returns sparse tensors $\Set{\Wfreq[\ell]_\text{PR}}$.
The \emph{re-training} step fine-tunes the non-zeros values of $\Set{\Wfreq[\ell]_\text{PR}}$ to recover accuracy. The final outputs are the sparse spectral kernels $\Set{\Wfreq[\ell]_\alpha}$. 
It is worth noting that, the ADMM training can be understood as solving a constrained optimization problem, where the optimization objective is loss minimization to ensure high classification accuracy, and the constraint is imposed by the pruning factor $\alpha$.
Due to the non-convexity in the objective function and the non-differentiability in the constraint, it is hard to identify analytical solutions. 
Thus, ADMM breaks the optimization problem into two sub-problems, where one of them has a closed-form solution and the other can be solved iteratively by the stochastic gradient descent algorithm. 
The ADMM training alternates between the two sub-problems, and eventually they will both converge. 
The solution identified by ADMM ensures high accuracy, and can be easily pruned. 

Different from the existing ADMM based approach \cite{zhang2018systematic} which performs pruning in the space domain, {\spec} performs end-to-end spectral pruning without transformation between $\W$ and $\Wfreq$.
Such spectral pruning enables us to exploit computation redundancy in both the sliding window operation and the spectral kernel weights. Therefore, together with our novel FPGA architecture (Section \ref{sec: arch}), {\spec} achieves remarkably high inference throughput without accuracy loss. 


\subsection{ADMM Training}
\label{sec: admm}

To prune an $L$-layer CNN with pruning factor $\alpha$, we abstract the problem as the below optimization problem:


\begin{equation*}
\begin{aligned}
    & \underset{\Set{\Wfreq^\ell_*}}{\text{minimize}}
    & & \loss\paren{\bm{L},\overline{\bm{L}};\Set{\Wfreq^\ell_*}} \\
    & \text{subject to}
    & & \nnz\paren{\Wfreq^\ell_*}\leq \frac{1}{\alpha}\cdot \sz\paren{\Wfreq^\ell_*},\quad \forall 1\leq \ell\leq L.
\end{aligned}
\end{equation*}

\noindent where $\loss\paren{\cdot}$ gives the CNN loss calculated by the predicted labels $\bm{L}$ and the ground-truth $\overline{\bm{L}}$, under the current model parameters $\Set{\Wfreq^\ell_*}$.
Function $\nnz\paren{\cdot}$ measures the \textbf{n}umber of \textbf{n}on-\textbf{z}eros of the tensor, and $\sz\paren{\cdot}$ returns the total number of elements of the tensor (i.e., $\sz\paren{\Wfreq}=\fin\cdot\fout\cdot n^2$).

The above problem is hard to solve since the constraint is non-differentiable. Therefore, we perform several steps of transformation to make the problem tractable. Firstly, we rewrite the above optimization problem with an auxiliary variable $\Zfreq[\ell]_*$ and an indicator function $\ind\paren{\cdot}$:

\begin{equation}
\label{eq: admm dfn}
\begin{aligned}
    & \underset{\Set{\Wfreq[\ell]_*}}{\text{minimize}} & & \loss\paren{\bm{L},\overline{\bm{L}};\Set{\Wfreq[\ell]_*}} + \sum \limits_{\ell=1}^{L}\ind\paren{\Zfreq[\ell]_*,\alpha}\\
    & \text{subject to} & & \Zfreq[\ell]_*=\Wfreq[\ell]_*, \quad \forall 1\leq \ell \leq L. 
\end{aligned}
\end{equation}

\noindent where $\ind\paren{\Zfreq[\ell]_*,\alpha}=\begin{cases}0, & \text{if }\nnz\paren{\Zfreq[\ell]_*}\leq \frac{1}{\alpha} \sz\paren{\Wfreq[\ell]_*}\\+\infty, & \text{otherwise} \end{cases}$ 

The indicator function, seen as enforcing hard penalty on the spectral kernel sparsity, still makes the optimization difficult. Therefore, as the next step of transformation, we relax the simple constraint of Equation \ref{eq: admm dfn}.
During the optimization process, we do not enforce $\Zfreq[\ell]_*= \Wfreq[\ell]_*$. Instead, we apply to the objective function an additional penalty term $\pnty\paren{\cdot}$ measuring the distance between $\Zfreq[\ell]_*$ and $\Wfreq[\ell]_*$. 
Then the optimization problem (on $\Wfreq[\ell]_*$ and $\Zfreq[\ell]_*$) becomes unconstrained, and thus, can be solved iteratively.
The values of $\Wfreq[\ell]_*$ and $\Zfreq[\ell]_*$ are updated in each iteration, and ultimately, they converge to the same value (i.e., $\Wfreq[\ell]_*\rightarrow \Zfreq[\ell]_*$) which is regarded as a good solution to the objective. 
Theoretically, penalty $\pnty\paren{\cdot}$ should grow with the number of iterations to ensure convergence.  

With the above intuitions, we formally describe the algorithm to solve Equation \ref{eq: admm dfn} under ADMM. 
Let $\Ufreq[\ell]_*$ measure the difference between $\Wfreq[\ell]_*$ and $\Zfreq[\ell]_*$. Define function
$f_\text{AD}\coloneqq\loss\paren{\bm{L},\overline{\bm{L}};\Set{\Wfreq[\ell]_*}}+\sum\limits_{\ell=1}^{L}\ind\paren{\Zfreq[\ell]_*,\alpha}+\sum\limits_{\ell=1}^{L}\pnty\paren{\Wfreq[\ell]_*,\Zfreq[\ell]_*,\Ufreq[\ell]_*}$.
ADMM decomposes the optimization on $f_\text{AD}$ into two sub-problems, one on $\Wfreq[\ell]_*$ only and the other on $\Zfreq[\ell]_*$. ADMM finds a solution by ``alternating the optimization direction'' along $\Wfreq[\ell]_*$ and $\Zfreq[\ell]_*$.  
The update rule for $\Wfreq[\ell]_*$ and $\Zfreq[\ell]_*$ at iteration $i+1$ is defined as Equation \ref{eq: admm 3 step}.

Now we use the Frobenius norm as the measure of matrix distance. Thus, $\pnty\paren{\Wfreq,\Zfreq,\Ufreq}\coloneqq \frac{\rho}{2}\norm{\Wfreq-\Zfreq+\Ufreq}_F^2 - \frac{\rho}{2}\norm{\Ufreq}_F^2$, where $\rho$ is a given constant coefficient.  The first sub-problem of Equation \ref{eq: admm 3 step} often does not have an analytic solution (due to non-convexity of $\loss\paren{\cdot}$), but can be solved by stochastic gradient descent since both $\loss\paren{\cdot}$ and $\pnty\paren{\cdot}$ are differentiable. 

In fact, the objective of ${\loss}\paren{\Set{\Wfreq[\ell]}}+\sum \pnty\paren{\Wfreq[\ell],\Zfreq[\ell],\Ufreq[\ell]}$ can be understood as the standard CNN loss plus an additional regularization term, and so existing techniques on CNN training are all applicable here. 
The second sub-problem admits a simple analytic solution. Clearly, $\Zfreq[\ell]_{i+1}$ should satisfy the sparsity constraint by $\alpha$ (so that $\ind\paren{\Zfreq[\ell]_{i+1},\alpha}=0$). Then, due to the special form of the penalty (i.e., Frobenius norm), to get optimal $\Zfreq[\ell]_{i+1}$, we take $\Wfreq[\ell]_{i+1}+\Ufreq[\ell]_{i}$ and set the elements of smallest magnitude to zero, until $\frac{1}{\alpha}\cdot \sz\paren{\Zfreq[\ell]_{i+1}}$ non-zeros remain.
As for the variable $\Ufreq$, note that it records the cumulative difference between $\Wfreq$ and $\Zfreq$, thus enforcing $\Wfreq$ to eventually converge to $\Zfreq$.

\begin{subequations}
\label{eq: admm 3 step}
\begin{align}
    \Set{\Wfreq[\ell]_{i+1}} =& \argmin\limits_{\Set{\Wfreq[\ell]_*}} f_\text{AD} 
    = \argmin\limits_{\Set{\Wfreq[\ell]_*}}\Bigg(\loss\paren{\bm{L},\overline{\bm{L}};\Set{\Wfreq[\ell]_*}} \nonumber\\
    &\qquad\qquad+\sum\limits_{\ell=1}^{L}\pnty\paren{\Wfreq[\ell]_*,\Zfreq[\ell]_i,\Ufreq[\ell]_i}\Bigg)\\
    \Set{\Zfreq[\ell]_{i+1}} =& \argmin\limits_{\Set{\Zfreq[\ell]_*}} f_\text{AD}= \argmin\limits_{\Set{\Zfreq[\ell]_*}}\Bigg(\sum\limits_{\ell=1}^{L}\ind\paren{\Zfreq[\ell]_*,\alpha}\nonumber\\
    &\qquad\qquad+\sum\limits_{\ell=1}^{L}\pnty\paren{\Wfreq[\ell]_{i+1},\Zfreq[\ell]_*,\Ufreq[\ell]_i}\Bigg)\\
    \Set{\Ufreq[\ell]_{i+1}} =& \Set{\Ufreq[\ell]_i+\Wfreq[\ell]_{i+1}-\Zfreq[\ell]_{i+1}}
\end{align}
\end{subequations}

In summary, we iteratively update $\Wfreq$ and $\Zfreq$ in an ``alternating direction'' manner. Throughout all iterations, $\nnz\paren{\Zfreq}\leq \frac{1}{\alpha}\cdot \sz\paren{\Zfreq}$ is always ensured, while $\nnz\paren{\Wfreq}$ may not satisfy the $\alpha$ constraint. 
At the end of the ADMM training, $\paren{1-\frac{1}{\alpha}}$ fraction of the elements in $\Wfreq[\ell]_\text{AD}$ are close to (but not exactly equal to) zero.
We then prune those near-zero elements and fine-tune the remaining $\frac{1}{\alpha}$ fraction of elements, by our spectral re-training algorithm.



\subsection{Pruning and Re-Training}
\label{sec: retrain}
After the ADMM optimization, we remove the $\paren{1-\frac{1}{\alpha}}$ fraction of the near-zero elements of $\Wfreq[\ell]_\text{AD}$, to get sparse kernels $\Wfreq[\ell]_\text{PR}$. 
Such pruning operation results in slight accuracy degradation ($0.03\%$ for LeNet on MNIST and $0.4\%$ for VGG16 on CIFAR-10).
To recover accuracy, we further tune the non-zero values of $\Wfreq[\ell]_\text{PR}$, without changing the kernel sparsity. We term the fine-tuning step as spectral re-training. 

Since re-training only aims at improving accuracy, we define loss function simply as $f_\text{RT}=\loss\paren{\bm{L},\overline{\bm{L}};\Set{\Wfreq[\ell]_\text{PR}}}$. 
Thus, using stochastic gradient descent, re-training propagates gradients from $f_\text{RT}$ backwards and adjusts the values of $\Wfreq[\ell]_\text{PR}$ accordingly. 
Note that since re-training does not change the sparsity of $\Wfreq[\ell]_\text{PR}$, the gradient $\frac{\partial f_\text{RT}}{\partial \Wfreq[\ell]_\text{PR}}$ is masked by the non-zero positions of $\Wfreq[\ell]_\text{PR}$. In summary, at the end of re-training, we have $\nnz\paren{\Wfreq[\ell]_\alpha}\leq \frac{1}{\alpha}\cdot \sz\paren{\Wfreq[\ell]_\alpha}$, and the sparse spectral CNN achieves high classification accuracy ($99.1\%$ for LeNet on MNIST and $90.8\%$ for VGG16 on CIFAR-10 when $\alpha=4$).

\begin{algorithm}
\caption{{\spec} spectral pruning algorithm}
\label{algo: prune overall}
\begin{algorithmic}[1]
\renewcommand{\algorithmicrequire}{\textbf{Input:}}
\renewcommand{\algorithmicensure}{\textbf{Output:}}
\Require Unpruned spectral CNN $\Set{\Wfreq[\ell]}$; Pruning rate $\alpha$; Learning rate $\eta$, $\eta'$; Penalty coefficient $\rho$;
\Ensure Pruned spectral CNN with sparse kernels $\Set{\Wfreq[\ell]_\alpha}$
\State $\Set{\Wfreq[\ell]_0}\gets \Set{\Wfreq[\ell]}$
\State $\Set{\Zfreq[\ell]_0}\gets\Set{\bm{0}}$
\State $\Set{\Ufreq[\ell]_0}\gets \Set{\bm{0}}$
\For{\text{iteration} $i=0$ \text{to} $i_\text{max}$}{\color{blue}\Comment{ADMM training}}
    \While{\text{not converged}}
        \State $\Set{\Wfreq[\ell]_{i}}\gets \Set{\Wfreq[\ell]_{i}-\eta\cdot \frac{\partial f_\text{AD}}{\partial \Wfreq[\ell]_i}}$
    \EndWhile
    \State $\Set{\Wfreq[\ell]_{i+1}}\gets \Set{\Wfreq[\ell]_i}$
    \State $\Set{\Zfreq[\ell]_{i+1}}\gets \Set{\text{Pruned } \paren{\Wfreq[\ell]_{i+1}+\Ufreq[\ell]_i}}$
    \State $\Set{\Ufreq[\ell]_{i+1}}\gets \Set{\Ufreq[\ell]_i+\Wfreq[\ell]_{i+1}-\Zfreq[\ell]_{i+1}}$
\EndFor
\State $\Set{\Wfreq[\ell]_\text{AD}}\gets \Set{\Wfreq[\ell]_{i_\text{max}}}$
\State $\Set{\Wfreq[\ell]_\text{PR}}\gets \Set{\text{Pruned }\Wfreq[\ell]_{\text{AD}}}$ {\color{blue}\Comment{Pruning}}
\For{\text{iteration} $i=0$ \text{to} $i_\text{max}'$}  {\color{blue}\Comment{Re-training}}
    \State $\Set{\Wfreq[\ell]_\text{PR}}\gets \Set{\Wfreq[\ell]_\text{PR}-\eta'\cdot \text{\fontfamily{lmtt}\selectfont Mask}\paren{\frac{\partial f_\text{RT}}{\partial \Wfreq[\ell]_\text{PR}}}}$
\EndFor
\State $\Set{\Wfreq[\ell]_\alpha}\gets \Set{\Wfreq[\ell]_\text{PR}}$ {\color{blue}\Comment{End of {\spec} pruning}}
\end{algorithmic} 
\end{algorithm}

Algorithm \ref{algo: prune overall} shows the overall {\spec} pruning algorithm (including ADMM training, pruning and re-training), as described in above. 
As a final remark, to support the stochastic gradient descent algorithm for any FFT size $n$, we incorporate the OaA technique into ADMM training and re-training. 
During forward propagation, we partition the input activation of each layer into $n\times n$ tiles and apply $\Wfreq[\ell]_i$ or $\Wfreq[\ell]_\alpha$ according to the steps described in Section \ref{sec: back spec}. 
During backward propagation, the gradients are derived by the automatic differentiation algorithm of Tensorflow. 



%% file: fig/overview_prune.tex
\colorlet{darkgreen}{black!60!green}
\begin{tikzpicture}[
outer_dashed/.style={draw=gray,dashed,fill=green!1,thick},
node_spec/.style={draw=black!60!green,fill=green!10,thick,align=center},
node_inout/.style={align=center},
node_inner/.style={align=center,color=black!60!white},
node_param/.style={align=center,color=black!60!green},
>={Stealth[inset=0pt,length=4pt,angle'=45,round]}]

\node[node_inout] (w orig) at (0,1.4) {$\Set{\Wfreq[\ell]}$};

\node[node_spec] (admm) at (0,0) {ADMM training};

\node[node_inner] (w admm) at (0,-1.2) {$\Set{\Wfreq[\ell]_{\text{AD}}}$};

\node[node_spec] (prune) at (0,-2.4) {Pruning};

\node[node_inner] (w prune) at (0,-3.6) {$\Set{\Wfreq[\ell]_{\text{PR}}}$};

\node[node_spec] (retrain) at (0,-4.8) {Re-training};

\node[node_inout] (w retrain) at (0,-6.2) {$\Set{\Wfreq[\ell]_\alpha}$};

\node[node_param] (alpha) at (-3,-1) {$\alpha$};
\node[node_param] (rho) at (2,0) {$\rho$, $\eta$};
\node[node_param] (eta2) at (2,-4.8) {$\eta'$};

\begin{pgfonlayer}{background}
\node[inner sep=10pt,outer_dashed,fit=(admm) (prune) (retrain) (rho) (eta2)] (spec prune) {};
\end{pgfonlayer}

\draw[->,draw=black] (w orig) -- (admm);
\draw[->,draw=black] (admm) -- (w admm);
\draw[->,draw=black] (w admm) -- (prune);
\draw[->,draw=black] (prune) -- (w prune);
\draw[->,draw=black] (w prune) -- (retrain);
\draw[->,draw=black] (retrain) -- (w retrain);
\draw[->,black!60!green] (alpha) -| (-2.5,0) |- (admm);
\draw[->,black!60!green] (alpha) -| (-2.5,-1.2) |- (prune);
\draw[->,black!60!green] (rho) -- (admm);
\draw[->,black!60!green] (eta2) -- (retrain);
\end{tikzpicture}

%% file: 4_arch.tex
\section{Accelerator Design}
\label{sec: arch}


\subsection{Overview}

\begin{figure*}[!htb]
    \centering
    \includegraphics[width=0.8\textwidth]{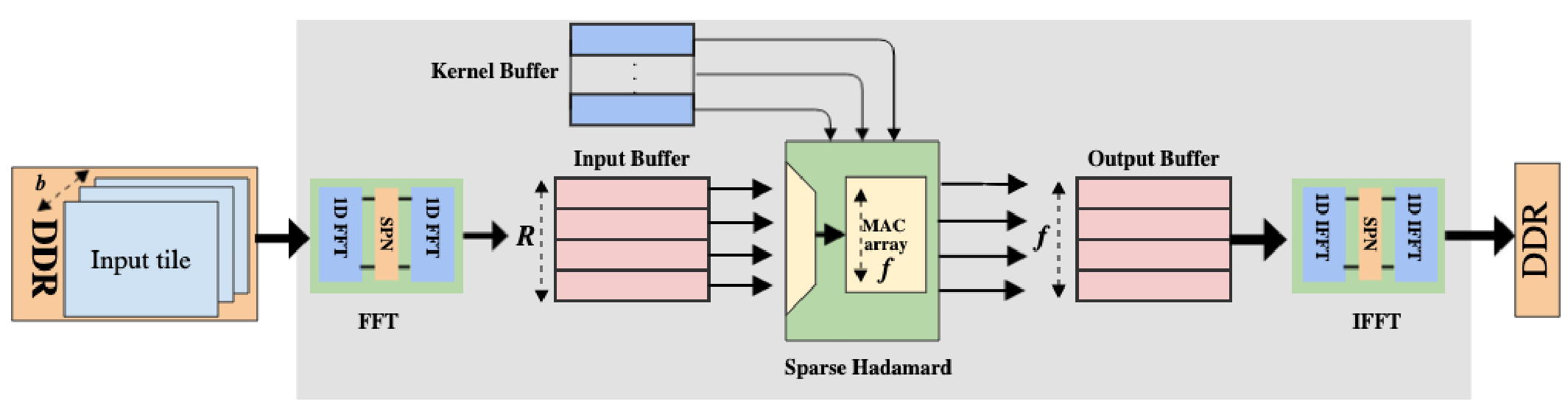}
    \caption{Overview of {\spec} inference engine}
    \vspace{-.4cm}
    \label{fig: arch overview}
\end{figure*}

By Equation \ref{eq: spectral conv}, computation of sparse spectral convolution can be decomposed into three steps: 
\begin{enumerate*}[(1)]
\item 2D FFT on the input activations;
\item Multiplication-Accumulation (MAC) for Hadamard product computation and reduction along input channel dimension;
\item 2D IFFT on output activations.
\end{enumerate*}
Figure \ref{fig: arch overview} shows the overall accelerator to compute the above steps. 

The external DDR stores the input images, layer activations, as well as the sparse spectral kernel weights of the CNN model. Due to the limited on-chip BRAM, we perform tiling of the activations and kernels. A tile of spectral kernel is pre-loaded into the \textbf{kernel buffer}. Tiles of layer activations communicate back and forth between DDR and FPGA to keep the inference pipeline busy. 
Once a tile of previous-layer activations arrives on-chip, it goes through a \textbf{2D FFT} module and the spectral outputs are stored in the \textbf{input buffer}. 
The \textbf{sparse Hadamard} module then computes Hadamard product of the \emph{sparse} spectral kernels and the \emph{dense} spectral activations, by reading into the kernel and input buffers. The partial results of accumulation along the input channel dimension are stored in the \textbf{output buffer}. 
After iterating through all data in the kernel tile, the output activations go through a \textbf{2D IFFT} module and return to DDR. 
We implement 2D FFT and IFFT based on \cite{Hanqing_FPGA18}. i.e., we decompose the 2D FFT into two phases of 1D FFTs. We use the 1D FFT pipeline and perform matrix transpose with Streaming Permutation Network proposed in \cite{ren_spn}. 

The following is required to achieve high performance of the inference engine:
\textbf{(i) Efficient random access} into the data buffers: The sparse Hadamard module fetches data from the dense activation tensor based on the non-zero indices of the pruned kernels. 
Since kernels can have random non-zero indices, bank conflicts may happen if multiple MAC pipelines of the Hadamard module request data from the same activation BRAM block.
Therefore, we replicate activation data to reduce the pipeline stalls due to BRAM conflicts.
 \textbf{(ii) Parallelism} across various  tensor dimensions: Spectral convolution with pruning dramatically reduces the CNN computation complexity. We design a parallelization strategy that translates the reduced amount of computation into higher throughput. We parallelize along the batch and channel dimensions of activations, while ensuring fast random access of on-chip data.

\subsection{Sparse Spectral Convolution Engine}
\label{sec: arch sparse conv}

\begin{figure*}[!htb]
    \centering
    \begin{subfigure}[b]{0.2\textwidth}
         \centering
         \includegraphics[width=\textwidth]{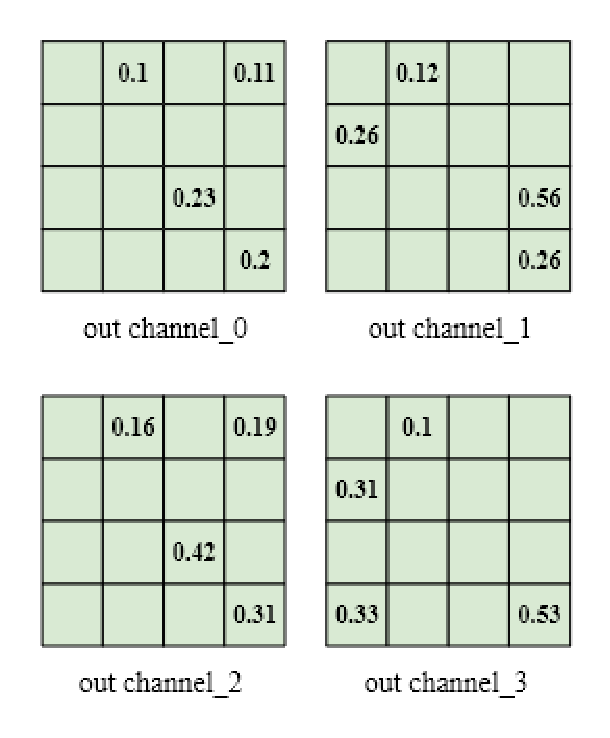}
         \caption{Example sparse kernels}
         \label{fig: weightMem_a}
     \end{subfigure}
     \begin{subfigure}[b]{0.6\textwidth}
         \centering
         \includegraphics[width=\textwidth]{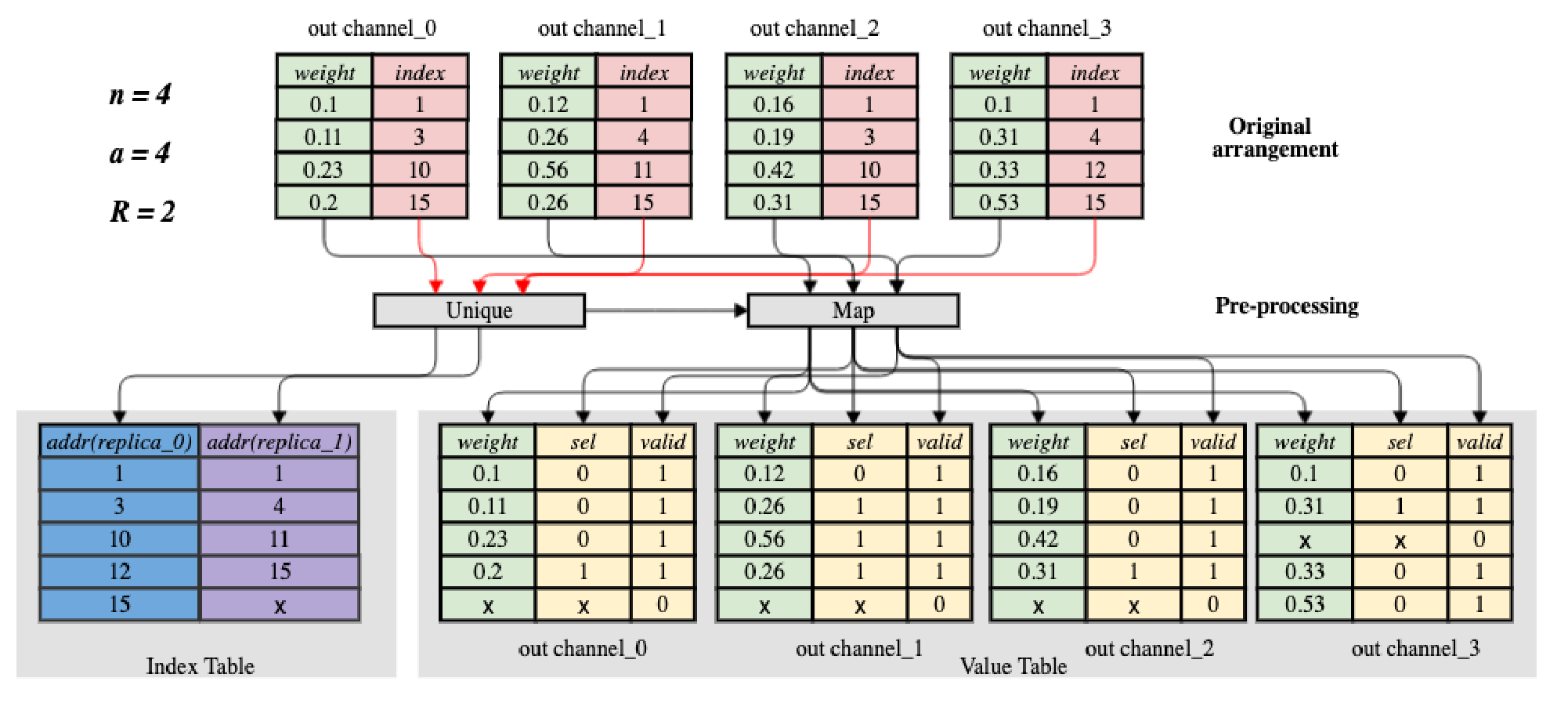}
         \caption{Data structure and memory layout}
         \label{fig: weightMem_b}
     \end{subfigure}
     \caption{Storage arrangement for sparse kernels}
     \label{fig: weightMem}
\end{figure*}

The sparse Hadamard module is the key component of the inference engine. 
This module processes a kernel tile of shape $c\times c\times n\times n$ and an activation tile of shape $b\times c\times n \times n$, where $c$ is the tiling factor along the input and output channel dimensions (i.e., the $\fin$ and $\fout$ channels are partitioned into tiles of $c$ channels); $n$ is the width and height of each kernel and activation map (i.e., 2D FFT size), and $b$ is the batch size. 
Note that pruning is along the last two dimensions of the kernel tile, so the pruning algorithm ensures \emph{exactly} $\frac{1}{\alpha}\cdot n^2$ non-zeros in each one of the $c^2$ kernel maps. 

Figure \ref{fig: weightMem} illustrates the storage arrangement for sparse kernels. As an example, Figure \ref{fig: weightMem_a} shows four sparse spectral kernels.
The upper part of Figure \ref{fig: weightMem_b} shows the data structure to represent the data in Figure \ref{fig: weightMem_a}. 
For each kernel matrix of input channel $i$ and output channel $j$, we use a $\paren{\frac{1}{\alpha}\cdot n^2} \times 2$ table to store the non-zero values and the corresponding indices. 
The lower part of Figure \ref{fig: weightMem_b} shows the memory layout. We split the value-index table into two tables and populate them with some pre-processed information to facilitate hardware access. 
The \emph{index table} is $\paren{\lambda\cdot\frac{1}{\alpha}\cdot n^2}$ by $R$, where $R\geq 1$ is the number of activation replicas and $\lambda\geq 1$ is an overhead coefficient related with $R$. This table stores pre-computed addresses into the input buffer. The \emph{value table} is $\paren{\lambda\cdot\frac{1}{\alpha}\cdot n^2}$ by $3$. This table stores the non-zero values and control signals required by the Hadamard pipeline. More details of the tables and parameters are discussed later in this section. 

\begin{figure*}[!htb]
    \centering
    \includegraphics[width=.64\textwidth]{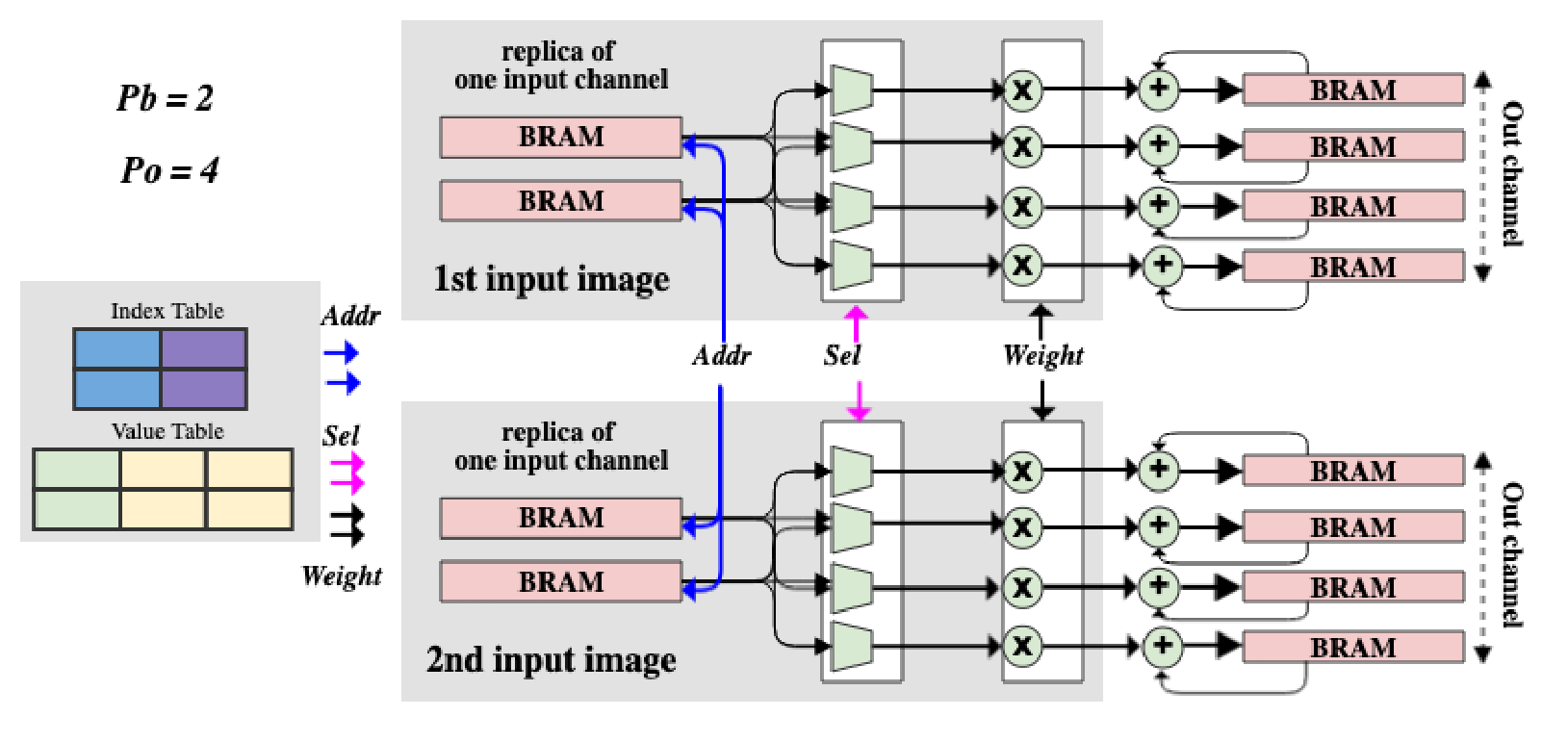}
    \caption{Sparse Hadamard pipeline}
    \label{fig:sparseHadamard}
\end{figure*}

To compute a single $\Xfreq_{k,i}\circ \Wfreq_{j,i}$ in Equation \ref{eq: spectral conv}, the sparse Hadamard module sequentially iterates through the index and value tables corresponding to indices $j$, $i$, and initiates random accesses into the input buffer. 
A single multiplier finishes the computation for a given $\paren{j,i}$ pair in $\lambda\cdot \frac{1}{\alpha}\cdot n^2$ cycles. 
All multipliers on-chip together exploit parallelism along the dimensions of batch and output channels. 
Figure \ref{fig:sparseHadamard} shows the design of the sparse Hadamard computation pipeline. 

Let $P=P_b\cdot P_o$ be the total number of multipliers in the module, where $P_b$ and $P_o$ denote batch and output channel parallelism, respectively ($P_b=2$, $P_o=4$ in Figure \ref{fig:sparseHadamard}). The $P_o$ multipliers corresponding to the same $P_b$ index form a \emph{group}, and they all access activations belonging to the same input channel. 
Thus, each cycle, a group initiates $P_o$ memory requests into the same map $\Xfreq_{k,i}$. 
Ideally, all the $P_o$ requests have to be served in one cycle to avoid pipeline stalls. 
In our design, since we store each $\Xfreq_{k,i}$ of size $n\times n$ into a single BRAM block, we then need $P_o$ activation replicas in case of the worst case scenario where all the memory requests are distinct. 
In practice, however, we notice that the spectral sparse kernels demonstrate strong correlation in non-zero locations across channels. Thus, the number of unique addresses in a group of $P_o$ requests is often much less than $P_o$. 
For this reason, we introduce the number of replicas $R$ (where $1\leq R\leq P_o$, and $R=2$ in the example) as an extra design parameter. Higher $R$ reduces the value of $\lambda$ and thus the likelihood of pipeline stalls, but also increases BRAM overhead due to replication. 

To route from the returned $R$ activation values to the $P_o$ multipliers, we implement a crossbar connecting BRAMs to multipliers for each group. An $R$-to-$1$ MUX is placed in front of each multiplier for data selection. 
We further utilize the information in the index and weight tables (Figure \ref{fig: weightMem_b}) to avoid runtime calculation of addresses and MUX control signals. Since non-zero locations of spectral kernels are fixed during inference, we can determine the unique addresses within each group of requests in offline pre-processing. Specifically, 
during pre-processing, we sort the group of $P_o$ requests and
store the $Q$ number of unique addresses in $\ceil*{\frac{Q}{R}}$ rows of the index table. 
As a result, the $P_o$ runtime requests can be served in $\ceil*{\frac{Q}{R}}$ cycles. 
We also pre-compute the corresponding MUX selection signals (i.e., replica ID) and store them in the ``\texttt{sel}'' column of the value table (Figure \ref{fig: weightMem_b}). 
The ``\texttt{valid}'' bit specifies if the value in the corresponding row is valid. Among the $\lambda\cdot \frac{1}{\alpha}\cdot n^2$ rows in the value table, only $\frac{1}{\alpha}\cdot n^2$ of them are valid.

\subsection{Performance Analysis}
\label{sec: arch model}

We use the notation $\mathrm{S}_{*}$ to denote the total amount of resources available: $\mathrm{S}_\text{BW}$ for external bandwidth (number of complex words per cycle); $\mathrm{S}_\text{DSP}$ for DSPs (number of complex multiplier-adders); $\mathrm{S}_\text{BRAM}$ for BRAMs (number of $36\times 1$K BRAM blocks).
Also, as a reasonable design choice, we set $P_o=c$, $P_b=b$. Overall throughput (number of multiplication / additions per unit time) is calculated as:

\begin{equation}\label{eq: Tsys}
    T_\text{sys} = \paren{\frac{1}{\overline{\lambda}_{R}}\cdot P_o\cdot P_b\cdot 2}\cdot \min\left\{1, \frac{1}{2}\cdot \frac{\mathrm{S}_\text{BW}}{\mathrm{S}_\text{BW}^\text{req}}\right\}\cdot \mathrm{F}
\end{equation}
where $\mathrm{S}_\text{BW}^\text{req}$ is the required bandwidth to support the DDR to 2D FFT / IFFT communication (Since spectral kernels are reused by large enough number of input tiles, the amortized cost of loading kernels from DDR to FPGA is zero, as analyzed in \cite{Hanqing_FPGA18}); $\frac{1}{\overline{\lambda}_{R}}$ specifies the average DSP utilization given $R$ replicas; the number "2" indicates multiplication and addition; and $\mathrm{F}$ is the working frequency.
In the design, the throughput of FFT, sparse Hadamard, and IFFT can be easily matched by adjusting data parallelism of the 2D FFT and IFFT modules. Hence, $\mathrm{S}_\text{BW}^\text{req}$ is equivalent to required bandwidth by the FFT module, i.e., $\mathrm{S}_\text{BW}^\text{req}=\frac{2P_b\cdot n^2}{\Omega_\text{H}}\cdot \text{F}$,
where $\Omega_\text{H}$ is number of clocks for finishing the corresponding Hadamard product on the tile.
Note that FFT and IFFT computations only consist of a small portion of the total convolution layer workload, as shown in \cite{Hanqing_FPGA18}. So under a reasonable assumption that there are sufficient logic resources, we implement the 2D FFT / IFFT pipelines with the on-chip logic, and the Multiplication-Accumulation units of the sparse Hadamard module with the DSPs. 
Thus, the throughput optimized design is obtained by solving the following:

\begin{equation}\label{eq: maxTsys}
\begin{aligned}
    &\underset{R,P_b,P_o}{\text{maximize}}& &T_\text{sys} \\
    &\text{subject to}& &P_b\cdot P_o \leq \mathrm{S}_\text{DSP},\\ 
    & & &P_b\cdot \paren{R+P_o}+ 1.5\cdot P_o \leq \mathrm{S}_\text{BRAM}
\end{aligned}
\end{equation}
where, the BRAM constraint arises from the number of 1K BRAM blocks for the input/output buffers and the kernel buffer. The number $1.5$\footnote{16-bit fix point complex kernels need 1 $36\times 1$K BRAM, the corresponding indices need another $18\times 1$k BRAM} in the BRAM constraint means the number of needed $36\times 1$K BRAM blocks to store kernel values and indices.


%% file: 5_exp.tex
\section{Experiments}
\label{sec: exp}

\subsection{Experimental Setup}
For pruning, we implement the ADMM and re-training algorithm by Tensorflow on NVIDIA P100 GPUs. We use LeNet (on MNIST dataset) and VGG16 (on CIFAR-10 and Flowers-102 \cite{Flowers102} datasets) for accuracy and throughput evaluation. As for the hardware implementation, we use Xilinx Virtex-7 XC7VX690T to accelerate convolution layers of the CNN. Other operations, such as ReLU, pooling, partitioning and concatenation, are performed in the host CPU, which communicates with FPGA using PCIe interface.
Due to the fact that operations in the host CPU account for a small portion ($<1\%$) of total computations and most of these operations can be parallelized, CPU execution will not be the bottleneck during inference.
We use 16-bit fix-point data format during inference. We use Vivado2018.3 to synthesize and implement the Verilog code. In the following section, to make fair comparison with state-of-the-art spatial CNN accelerators, ``image frames per second'' (FPS) is adopted to measure inference throughput.  

\subsection{Evaluation Classification Accuracy}


In this section, we evaluate the effect of pruning rate $\alpha$ and FFT size $n$ on the classification accuracy.
We first obtain the dense spectral kernels $\Set{\Wfreq[\ell]}$ by converting from the pre-trained spatial CNNs.
Then we perform pruning by ADMM and re-training to get the sparse $\Set{\Wfreq[\ell]_\alpha}$. Keeping $\alpha$ the same among different kernels works well in other ADMM based compression works\cite{zhang2018systematic}. Besides, the same compression ratio avoids load imbalance when multiple kernels are executed in parallel.
We try six pruning configurations, by setting FFT size $n=8$ or $16$ and compression rate $\alpha=2$, $4$ or $8$. For each configuration, we perform hyper-parameter tuning independently, by searching over the parameter space defined by: ADMM penalty coefficient $\rho\in\Set{0.001, 0.002, 0.005}$, initial learning rate $\eta_0\in\Set{0.001,0.0001}$, learning rate decaying factor (per 20 epochs\footnote{An epoch means a full traversal of the training set.}) $\gamma\in\Set{0.8}$. The ADMM training converges after 200 epochs for LeNet, and 300 epochs for VGG16. 
The re-training step converges after 10 epochs for LeNet and 30 for VGG16.

\begin{figure}
    \centering
    \includegraphics[width=0.5\textwidth]{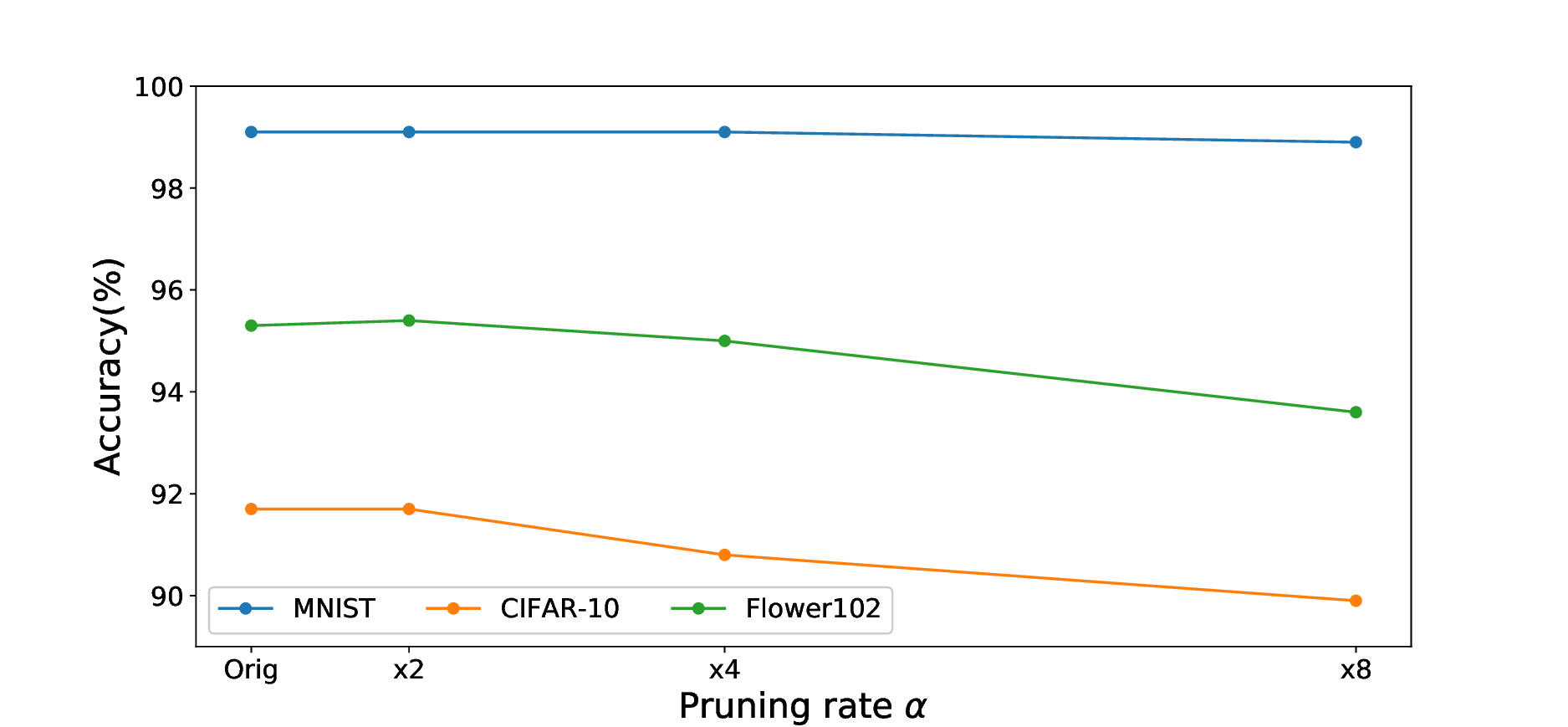}
    \caption{Accuracy under various pruning rate $\alpha$}
    \label{fig:acccompratio}
\end{figure}

\vspace{0.15cm}
\noindent\textbf{Accuracy under various pruning rates}\hspace{.3cm}
Figure~\ref{fig:acccompratio} shows accuracy of the three CNN models when increasing the pruning rate from $2$ up to $8$. 
For LeNet on MNIST, accuracy can be fully recovered even when $75\%$ spectral weights are pruned. Small accuracy degradation ($0.2\%$) is resulted from pruning $87.5\%$ of the spectral weights. For VGG16, 
our pruning algorithm achieves less than $0.9\%$ (for CIFAR-10) and $0.3\%$ (for Flowers-102) accuracy loss when pruning rate $\alpha\leq 4$. The accuracy loss is $1.8\%$ (for CIFAR-10) and $1.7\%$ (for Flowers-102) when $\alpha=8$. 

\begin{figure}
    \centering
    \includegraphics[width=0.5\textwidth]{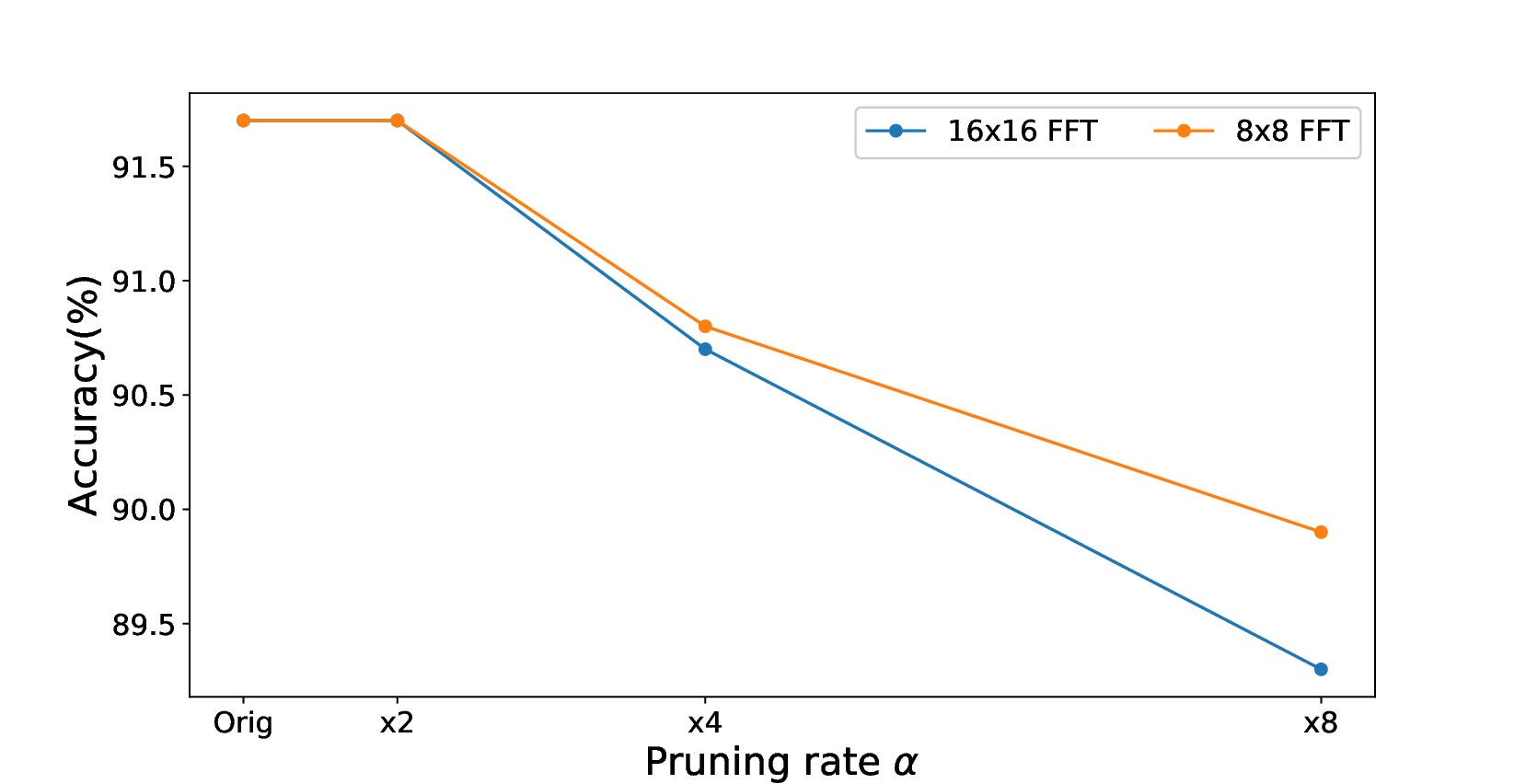}
    \caption{Accuracy under various FFT sizes $n$ (CIFAR-10)}
    \label{fig:accfftsize}
\end{figure}

\vspace{.15cm}\noindent\textbf{Accuracy under various FFT sizes}\hspace{.3cm}
It has been shown in \cite{Hanqing_FPGA18} that the FPGA computation and communication workload may depend on the selected FFT size $n$. 
Hence, it is necessary to evaluate the pruning algorithm performance under various $n$. 
As shown in Figure~\ref{fig:accfftsize}, $8\times 8$ spectral kernels leads to higher accuracy under the same pruning rate. As the pruning rate increases, the accuracy gap between the $8\times 8$ configuration and the $16\times 16$ one becomes larger.

\begin{figure*}
    \centering
    \includegraphics[width=0.8\textwidth]{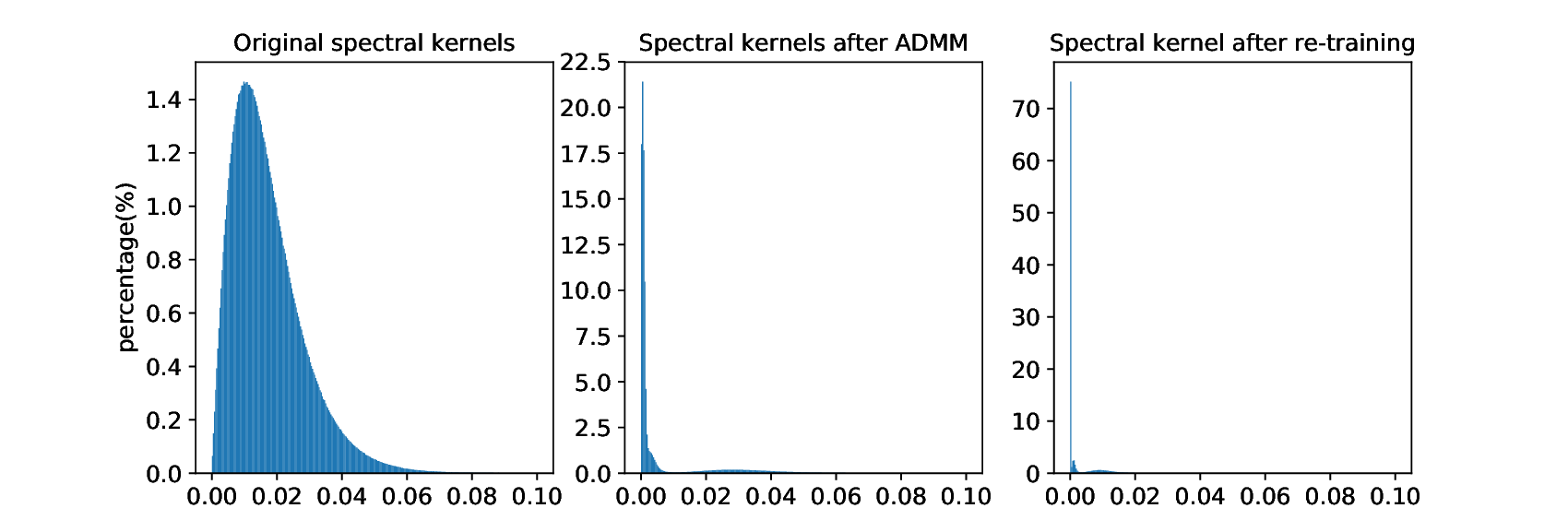}
    \caption{Value distribution of spectral kernels with $\alpha=4$}
    \label{fig:valdistx4}
\end{figure*}

\vspace{.15cm}\noindent\textbf{ADMM training statistics}\hspace{.3cm}
As described in Section \ref{sec: admm}, the ADMM pruning step drives $\paren{1-\frac{1}{\alpha}}$ portion of the $\Wfreq[\ell]$ elements to close-to-zero values. The re-training step sets these near-zero values to zero and fine-tune the remaining $\frac{1}{\alpha}$ portion. Here we visualize the value distributions of spectral kernels at various stages of the pruning, as shown in Figure~\ref{fig:valdistx4}. 
The left one represents the distribution of the original kernels $\Set{\Wfreq[\ell]}$ of VGG16, in which values spread out in the $\left[0.0,0.1\right]$ interval.  
After ADMM training with $\alpha=4$, the weight distribution is significantly changed, as most values gather around zero (see the middle figure). By further pruning near-zero weights and re-training, smaller percent ($25\%$) of the weights remain non-zero. 
We consistently observe such change in distribution for other CNNs and datasets.

Table \ref{tab: acc prune step} shows the change in classification accuracy after various steps of weight pruning ($\alpha=4$). 
After ADMM training, small accuracy drop is observed when testing under $\Set{\Wfreq[\ell]_{\text{AD}}}$. 
After obtaining $\Set{\Wfreq[\ell]_\text{PR}}$ by removing the near-zero values of $\Set{\Wfreq[\ell]_{\text{AD}}}$, the accuracy further drops for $0.03\%$ to $0.4\%$.
After the re-training step, the accuracy is recovered, and the final accuracy degradation compared with the original unpruned models is $0.0\%$ for MNIST, $0.9\%$ for CIFAR-10 and $0.3\%$ for Flowers-102.

\begin{table}[]
    \caption{Accuracy after various pruning steps ($\alpha=4$)}
    \label{tab: acc prune step}
    \centering
    \begin{tabular}{rccc}
    \toprule
     & MNIST & CIFAR-10 & Flowers-102\\
     \midrule
     \midrule
    Original $\Set{\Wfreq[\ell]}$ & 99.1 & 91.7 & 95.3 \\
    \midrule
    ADMM $\Set{\Wfreq[\ell]_\text{AD}}$ & 99.0 & 90.9 & 95.0\\
    Pruning $\Set{\Wfreq[\ell]_\text{PR}}$ & 99.0 & 90.5 & 94.8 \\
    Re-training $\Set{\Wfreq[\ell]_\alpha}$ & 99.1 & 90.8 & 95.0\\
    \bottomrule
    \end{tabular}
\end{table}


\subsection{Evaluation on Inference Throughput}

Overall inference throughput, bottlenecked by either computation or communication speed on FPGA, can be maximized by solving Equation \ref{eq: maxTsys}.
Note that to obtain the optimal design point $\paren{R^*, P_b^*, P_o^*}$, we have to empirically determine the relation between DSP utilization and the parameters $R$, $P_o$. 

\begin{figure*}
    \centering
    \includegraphics[width=0.8\textwidth]{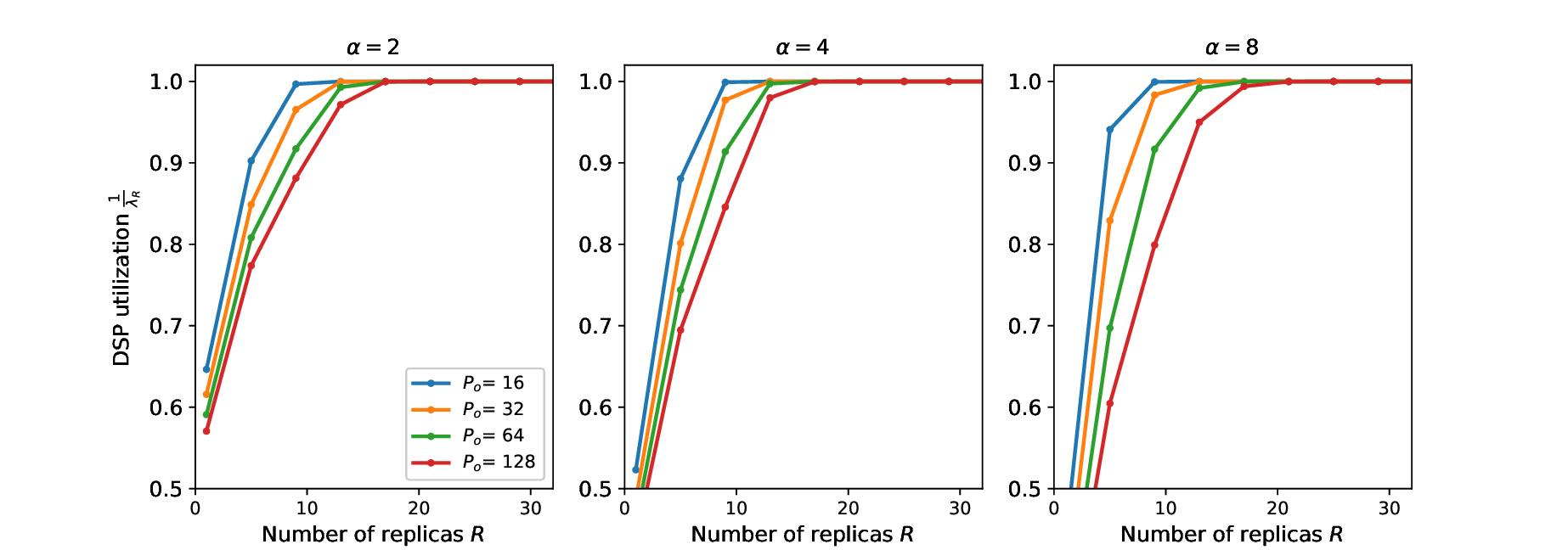}
    \caption{DSP utilization under various number of replica $R$, and output channel parallelism $P_o$}
    \label{fig:perf_ochnl}
\end{figure*}

Figure~\ref{fig:perf_ochnl} shows DSP utilization(average ratio of active DSP to total DSP) under various number of replicas $R$ and output channel parallelism $P_o$. 
In all cases, we do not require a large number of replicas to achieve high DSP utilization. 
Even when $P_o$ is as high as $128$, it suffices to set $R=10$ for over $80\%$ DSP utilization.
In addition, with increasing pruning rate, DSP utilization drops slightly under the same number of replicas. This is due to that the sparser the spectral kernels, the more irregular the non-zero locations. 

\begin{figure}
    \centering
    \includegraphics[width=0.4\textwidth]{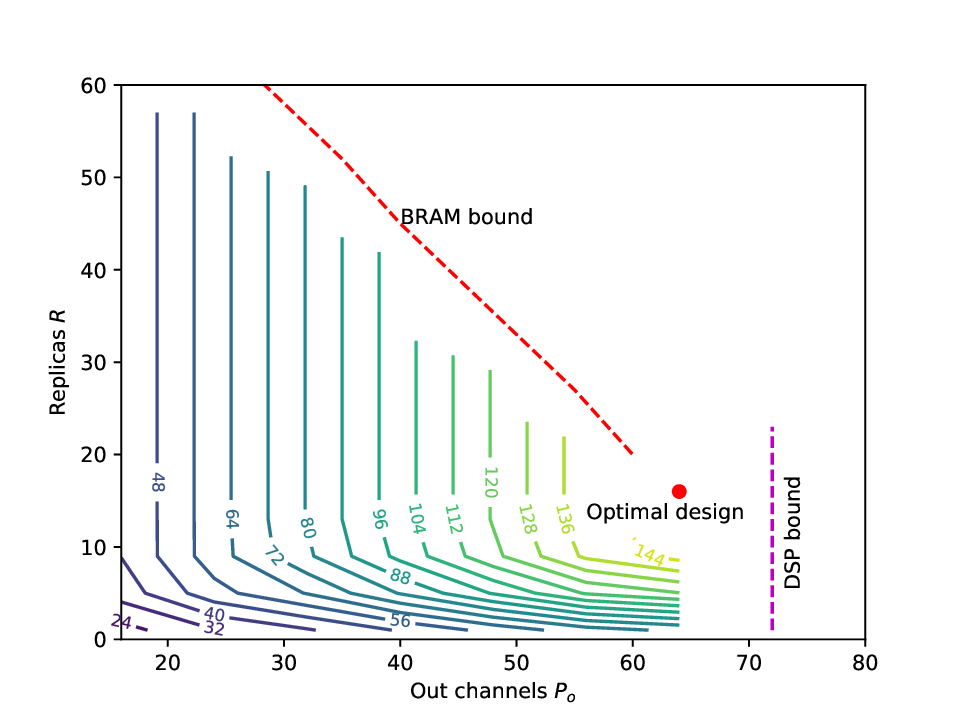}
    \caption{Throughput under various configurations}
    \label{fig:throughput}
\end{figure}

Based on the statistics of DSP utilization, overall throughput can be obtained by Equation \ref{eq: Tsys}, and optimal design point $\paren{R^*, P_b^*, P_o^*}$ can be obtained. Figure~\ref{fig:throughput} shows throughput under various configurations with $\alpha=4$, $n=8$ and $P_b=10$. 
The solid lines indicate throughput on the target hardware platform, and the dotted lines represent performance bound due to limited number on-chip BRAMs and DSPs. 
The solid lines terminate at $P_o=64$ since we set $P_o$ to be power of two. 
Due to enough DSP resources on the target platform, the design is bound by BRAMs.  
By Figure~\ref{fig:throughput}, the optimal configuration is $P_b^*=10$, $P_o^*=64$, $R^*=16$, corresponding to throughput of $148$ FPS (with average DSP utilization of $99\%$). The total required communication bandwidth between off-chip memory and on-chip processing pipeline is $9$ GB/s, which is lower than the peak system bandwidth of $21$ GB/s.

Given different compression ratios, we can also find optimal configurations based on design exploration algorithm, as shown in Table~\ref{tab: throughputVGG}. Throughput here is only for computation of convolutional layers. Although kernels are more sparse given higher $\alpha$, like $8$, throughput is still almost doubled compared with $\alpha=4$.

\begin{table}[]
    \caption{Throughput of VGG16 ($224\times 224$ input images)}
    \label{tab: throughputVGG}
    \centering
    \begin{tabular}{rccc}
    \toprule
    &\multicolumn{3}{c}{Pruning rate $\alpha$}\\
    \cmidrule(lr){2-4} & $2$ & $4$ & $8$ \\
     \midrule
     \midrule
        Throughput (FPS) & 74 & 148 & 284 \\
        Replicas $R$ & 16 & 16 & 16 \\
        DSP Utilization $\frac{1}{\overline{\lambda}_R}$ & 100\% & 99\% & 96\% \\
    \bottomrule
    \end{tabular}
\end{table}

\begin{table*}[!htb]
\caption{Comparison with state-of-the-art designs on VGG16}
\label{tab: compHardware}
\centering
\begin{tabular}{rccccccc}
\toprule
   & \cite{Hanqing_FPGA18} & \cite{Chi_FPGA17} & \cite{VGG_svd} & \cite{Caffein} & \cite{SystolicCNN} & \cite{sparcnet}& {\spec}\\
\midrule
\midrule
 \multirow{2}{*}{FPGA} & Intel & Intel & Xilinx  & Xilinx &  Intel & Xilinx & Xilinx \\
      & Stratix V & QPI FPGA & Zynq & Virtex-7 & Arria 10 & Artix-7 & Virtex-7 \\
 Frequency (MHz) & 200 & 200 & 150 & 150 & 221 & 100 & 200\\
 Datatype & 16-bit FX & 32-bit FT & 16-bit FX & 16-bit FX & 16-bit FX & 16-bit FT & 16-bit FX\\
 DSP Usage & 100\% (256)& 100\% (224) & 89\% (780) & 78\% (2833)& 49\% (1500) & 52\% (384) &  89\% (3200)\\
 LUT Usage & 46\% (107K)& - & 84\% (183K) & 81\% (300K)& 73\% (313K) & - & 55\% (237K)\\
 RAM Usage & 73\% (1377)& - & 87\% (486) & 42\% (624)&61\% (1668) & 53\% (388) & 85\% (1244)\\
 Throughput (fps) & 6 & 4 & 8 & 16 & 19 & 5 &\textbf{148} \\
\bottomrule
\end{tabular}
\\
    \vspace{.2cm}
    {\scriptsize{}{*} \textbf{FX}: Fixed-point data; \textbf{FT}: Floating-point data}{\scriptsize \par}
\end{table*}

Table~\ref{tab: compHardware} shows the throughput comparison with state-of-the-art on VGG16. {\cite{Hanqing_FPGA18,Chi_FPGA17}} accelerates unpruned spectral CNNs. 
{\spec} achieves $24\times$ higher throughput, where we estimate that $8\times$ improvement is due to the larger device and $3\times$ improvement comes from our spectral pruning. 
Compared with \cite{Caffein} which accelerates unpruned spatial CNNs on the same FPGA as us, {\spec} achieves $9\times$ improvement, where around $3\times$ is due to the spectral convolution algorithm and the other $3\times$ comes from spectral pruning. 
Compared with pruned spatial CNN implementation \cite{sparcnet}, {\spec} achieves $30\times$ higher throughput on a device with 2816 more DSPs. 
For a fairer comparison, assume throughput of \cite{sparcnet} scales with the number of DSPs and \cite{sparcnet} can be implemented on the same device as us with the same working frequency. In such case, {\spec} still achieves $1.8\times$ throughput improvement.

%% file: 6_conclusion.tex
\section{Conclusion}
\label{sec: conclusion}

We have proposed {\spec}, an approach to prune and accelerate spectral CNNs. 
{\spec} performs systematic weight pruning based on ADMM and deploys an efficient sparse tensor computation pipeline on FPGA to achieve high accuracy and high throughput inference simultaneously. 

In the future, we will extend {\spec} in the following ways. First of all, we will explore model redundancy from various aspects. We will incorporate channel pruning, and weight quantization into the ADMM framework to obtain even more compact spectral CNN models. 
We will also develop a complete tool chain to automatically prune, quantize and accelerate spectral CNNs. 

%% file: 7_ack.tex
\section{Acknowledgements}
This work was supported in part by National Science Foundation award number CNS-1643351.